\documentclass[letterpaper, 10 pt, conference]{ieeeconf}  

\IEEEoverridecommandlockouts                              

\usepackage{needspace}
\usepackage[normalem]{ulem}
\usepackage[utf8]{inputenc}
\usepackage[T1]{fontenc}
\usepackage[textsize=scriptsize]{todonotes}
\usepackage{lipsum}
\usepackage{algorithm}
\usepackage{algpseudocode}
\usepackage{url}
\usepackage{multirow}
\usepackage{graphicx}
\usepackage{blindtext}
\usepackage[separate-uncertainty = true, free-standing-units, space-before-unit, use-xspace]{siunitx}
\usepackage{color,soul}
\usepackage{svg}
\usepackage{amsmath} 
\usepackage{amsfonts}
\usepackage{amssymb}  
\usepackage{cleveref}

\usepackage{amsmath,amssymb,amsfonts}
\usepackage{graphicx}
\usepackage{xcolor}
\usepackage{ulem}

\usepackage{ifthen}
\newboolean{showcomments}
\setboolean{showcomments}{true} 
\ifthenelse{\boolean{showcomments}}
  {\newcommand{\nb}[2]{
    \fcolorbox{gray}{yellow}{\bfseries\sffamily\scriptsize#1}
    {\sf\small$\blacktriangleright${#2}$\blacktriangleleft$}
  }
  
  }
  {\newcommand{\nb}[2]{}
  
  }

\title{\LARGE \bf

Adaptable Recovery Behaviors in Robotics: A Behavior Trees and Motion Generators~(BTMG) Approach for Failure Management}

\author{Faseeh Ahmad$^{1}$, Matthias Mayr$^{1}$, Sulthan Suresh-Fazeela$^{2}$
and Volker Krueger$^{1}$
	\thanks{$^{1}$Department of Computer Science, Faculty of Engineering (LTH), Lund University, SE~221~00 Lund, Sweden. E-mail: <firstname>.<lastname>@cs.lth.se.
	}%
 	\thanks{$^{2}$Department of Computer Science, Faculty of Engineering (LTH), Lund University, SE~221~00 Lund, Sweden. E-mail: sulthansf95@gmail.com.
	}%
}

\begin{document}

\maketitle
\thispagestyle{empty}
\pagestyle{empty}

\begin{abstract}
In dynamic operational environments, particularly in collaborative robotics, the inevitability of failures necessitates robust and adaptable recovery strategies. Traditional automated recovery strategies, while effective for predefined scenarios, often lack the flexibility required for on-the-fly task management and adaptation to expected failures. Addressing this gap, we propose a novel approach that models recovery behaviors as adaptable robotic skills, leveraging the Behavior Trees and Motion Generators~(BTMG) framework for policy representation. This approach distinguishes itself by employing reinforcement learning~(RL) to dynamically refine recovery behavior parameters, enabling a tailored response to a wide array of failure scenarios with minimal human intervention. We assess our methodology through a series of progressively challenging scenarios within a peg-in-a-hole task, demonstrating the approach's effectiveness in enhancing operational efficiency and task success rates in collaborative robotics settings. We validate our approach using a dual-arm KUKA robot.
\end{abstract}

\section{Introduction}




In dynamic operational environments, ensuring the efficiency and adaptability of collaborative robots is crucial. Unlike traditional manufacturing line robots, collaborative robots are designed for on-the-fly task deployment, facing a unique set of challenges and failures. For instance, in a piston engine assembly process~\cite{rovida18btmg}, common issues such as misalignment of the engine block, obstruction by misplaced tools, and incorrect piston orientation due to handling errors can lead to substantial production delays. Addressing these failures promptly and effectively is crucial for maintaining seamless operations and efficiency.

Current strategies for managing these failures include human intervention, systematic failure analysis~\cite{jusuf2021review}, and automated recovery strategies~\cite{lei2023artificial, alves2020secure}. Human intervention relies on operator expertise for problem-solving. Failure analysis systematically identifies root causes to prevent future issues, but it requires time and expertise. Automated recovery strategies, on the other hand, leverage intelligent systems for quick detection and correction of failures, significantly reducing downtime and enhancing consistency. However, these strategies come with high initial costs and complexity in integration. Automated recovery strategies often rely on predefined scenarios and responses, lacking the flexibility to adapt to different variations of the expected failures~\cite{lei2023artificial}. While they can efficiently address a range of anticipated problems, their effectiveness diminishes where adaptability and rapid response to novel challenges are to be addressed.

\begin{figure}
    \centering
        \includegraphics[width=\columnwidth]{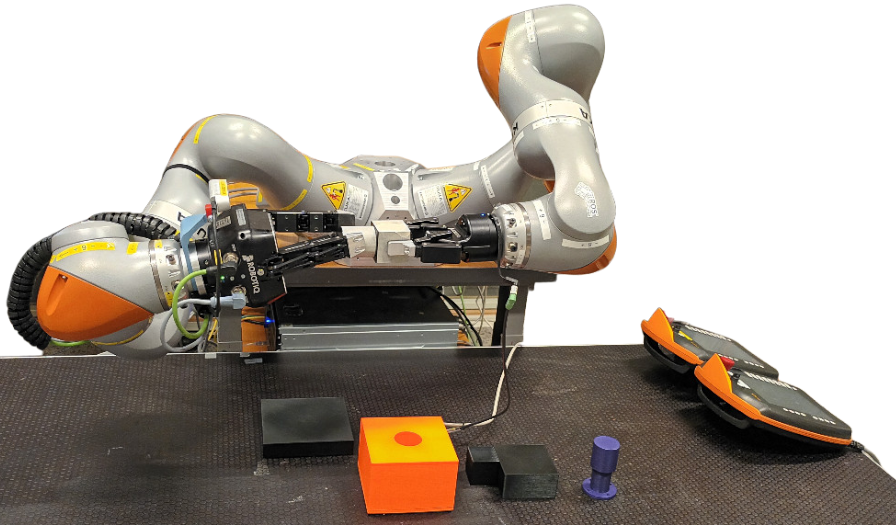}
        \caption{The real dual-arm \textit{KUKA iiwa} setup executing a handover task before inserting the peg in the orange block. On the table, the purple peg for Scenario~1, alongside various obstacles that block the opening an be seen.}
    \label{fig:Robot-Setup}
    \vspace{-0.1cm}
\end{figure}

The necessity for adaptive recovery behaviors becomes evident in collaborative tasks like piston engine assembly, where the nature of an obstruction dictates the required recovery strategy. Whether it involves removing a small obstacle or applying force to displace a larger one, the recovery behavior must be flexible enough to adjust its approach based on the specific challenge at hand. In this work, we propose a hybrid approach that combines the strengths of human expertise with the dynamic adaptability of recovery behaviors. Unlike traditional automated strategies that rely on predefined responses, our method distinguishes itself by leveraging RL to dynamically refine and adjust the parameters of recovery behaviors. This adaptability, enabled by RL, ensures that our robots can effectively handle a wider range of failure scenarios with minimal human intervention, making the recovery process more efficient and responsive to the task requirements. By focusing on adaptable recovery behaviors, our approach aims to equip robots with the ability to autonomously address failures, thereby enhancing adaptability and reducing response times.

Inspired by the literature, one effective way to enhance robot capabilities involves the adoption of skills~\cite{pedersen2016robot}. These robot skills, defined by specific parameters, preconditions, and postconditions~\cite{rovida2017skiros,mayr2023skiros2}, enable interaction with the environment in a robust, adaptable, and flexible manner, mirroring the qualities we seek for effective failure recovery. Building on this foundation, we suggest to model recovery behaviors as dedicated robotic skills, distinct from standard production skills and specifically designed to address failures, see Figure~\ref{fig:Approach}. This strategy equips robots with a specialized toolkit for managing known failure scenarios, streamlining the skill set for simplicity and reducing maintenance requirements.  For example, in piston engine assembly, alongside the primary assembly skill, we could have a recovery skill like 'pick-place' to remove obstacles such as misplaced tools blocking the engine hole.

Various execution strategies for skills have been explored in literature~\cite{herzig2016hierarchical,iovino2023programming}. We have chosen the Behavior Trees and Motion Generators~(BTMG) framework for its strengths in robustness, modularity, interpretability, and reactivity, which are good for effectively handling complex robotic tasks. This framework serves as the foundation for representing both standard operational skills and recovery behaviors.

In this paper, we extend the BTMG representation with the introduction of adaptable recovery behaviors inside the representation, drawing on insights from Styrud et al.~\cite{styrud2023bebop}. The parameters of these recovery behaviors can be set manually, through reasoning, or using~RL, offering flexibility based on the complexity of the task. We evaluate this approach with the peg-in-a-hole task, by gradually introducing failures to increase task complexity and demonstrate the necessity for sophisticated recovery behaviors. Following are the main contributions of the paper:
\begin{itemize}
    \item We extend the BTMG policy representation with the introduction of adaptable recovery behaviors, incorporating RL to dynamically adjust behavior parameters in response to task requirements.
    \item We present challenging scenarios as failure cases and adapt the parameters of recovery behaviors to suit the specific requirements of the task.
    \item We evaluate the performance of the proposed method on a peg-in-a-hole task using dual arm KUKA iiwa robot, demonstrating its effectiveness in adapting to and recovering from failures. .
\end{itemize}
\section{Related Work}


The development of recovery behaviors to mitigate failures and disturbances has emerged as a critical area of research, aiming to enhance the resilience and autonomy of robotic systems. In~\cite{council2020recovery}, an approach was introduced for the recovery of high-degree-of-freedom motor behaviors in robots, utilizing rank-ordered differential constraints to quickly adapt to damage, malfunction, or environmental changes.~\cite{lee2019robust} uses Deep Reinforcement Learning to control recovery maneuvers for quadrupedal robots, showcasing dynamic and reactive behaviors that enable a robot to recover from falls with a high success rate.~\cite{wu2018recovering} developed a state-dependent recovery policy that allows robots to recover from external disturbances across various tasks and conditions. In~\cite{gu2022reactive}, the authors explored push recovery for bipedal robot locomotion, integrating decision-making and motion planning to address perturbations. Lastly,~\cite{koos2013fast} proposes the T-resilience algorithm, a novel method that enables robots to autonomously discover compensatory behaviors in unanticipated situations, thereby facilitating fast damage recovery.  

Furthermore, the utilization of recovery behaviors in behavior trees~(BT) has also emerged as a recent interesting area of research. ~\cite{iovino2020survey,ghzouli2023behavior} provides a detailed survey on behavior trees in the domain of robotics and AI, highlighting the usage of BT across a wide landscape.~\cite{wu2021automated} introduces a framework that integrates an execution generation tool, a learning module, and a recovery pipeline to facilitate error detection, diagnosis, and recovery, showcasing the potential of BT in managing complex error recovery processes with minimal human intervention. In the context of wheeled robots,~\cite{de2023autonomous} demonstrates a hierarchical online hybrid planner for autonomous navigation. They utilize BT in enabling wheeled-legged robots to autonomously navigate and recover from collisions or planner failures, emphasizing the framework's capability to handle dynamic challenges without human intervention. On a side note~\cite{pezzato2020active} introduced a direct extension to BT by adding a new leaf node type within the structure, aimed at specifying desired states rather than actions. This innovation demonstrates improved runtime adaptability and suggests a method for integrating recovery states directly into the BT structure, enhancing their flexibility in error recovery scenarios.~\cite{paxton2019representing} introduces a framework called Robust Logical-Dynamical Systems~(RLDS), which combines the advantages of BT with theoretical performance guarantees. RLDS exemplifies how BT can be extended to achieve robust, reactive behavior in dynamic environments, particularly in manipulation tasks.

Our work distinguishes itself from aforementioned studies by leveraging~RL to dynamically define the parameters of recovery behaviors, a novel approach that significantly enhances adaptability and effectiveness in failure management. This emphasis on RL-driven parameterization underscores its potential utility, a point we elaborate on through various failure scenarios in our experimental section, demonstrating the benefits of integrating RL into recovery strategies.

\section{Background}
In this section, we discuss the relevant concepts that serve as background knowledge for this paper.
\subsection{Behavior Trees}
Behavior Trees~(BT)~\cite{colledanchise17ac} are a hierarchical model for task planning and decision-making, widely known in robotics~\cite{iovino2020survey} for their modularity, flexibility, and clarity. Initially conceived for video game AI to simulate complex behaviors, BT have been effectively adapted for robotic tasks, enabling structured execution of actions from simple to complex decision-making processes. A BT structures as a directed acyclic graph, initiating execution from the root node and ticking at regular intervals to adapt dynamically to environmental changes. This execution involves traversing the tree based on control logic, evaluating conditions, executing actions, and applying decorators to modify outcomes~\cite{olsson2016behavior}. The nodes are executed only when they are ticked and return \textit{Success}, \textit{Failure} or \textit{Running}. \textit{Control flow nodes} are the non-leaf nodes that control the execution flow, with sequence~(logical AND) and selector~(logical OR) being the most common. These nodes are responsible for determining the order and conditions under which child nodes are executed. The leaf nodes are called \textit{execution} nodes and are further divided into \textit{action} and \textit{condition} nodes. \textit{Condition} nodes only return \textit{Success} or \textit{Failure} and are used to evaluate the robot's state or environment. \textit{Action} nodes execute the tasks, such as movement or manipulation, and return statuses indicating \textit{Success}, \textit{Failure} or \textit{Running}. Lastly, we have \textit{decorator} nodes that modify the behavior or outcome of their child nodes to meet specific criteria or constraints.

BT offer several advantages in robotics. The readable and hierarchical structure enhances modularity~\cite{colledanchise142iicirsa, biggar2022modularity}, allowing for easy modification and expansion of robotic behaviors. This modularity, combined with the clarity of the BT framework, simplifies understanding and debugging, making BT an attractive choice for complex robotic applications. Furthermore, BT also support reactivity~\cite{colledanchise2018behavior}, enabling robotic systems to dynamically respond to changes in the environment. 

\subsection{Behavior Trees and Motion Generators~(BTMG)}
Leveraging the foundational structure of Behavior Trees, Rovida et al.~\cite{rovida18btmg} integrate it with an arm motion generation strategy known as Motion Generators (MG) to develop the Behavior Trees and Motion Generators (BTMG) policy representation.~MG, as detailed in~\cite{rovida18btmg}, employs an impedance controller to control the robot's end-effectors in Cartesian space. This approach not only enables the execution of the primary motion but also permits the superimposition of additional motions through a generic varying Cartesian wrench. Furthermore, MG incorporates mechanisms for constraining velocities, accelerations, and torques, thereby addressing safety requirements comprehensively. For an in-depth exploration of MG refer to~\cite{rovida18btmg}. The addition of MG to the BT structure is done via having \textit{action} and \textit{condition} nodes specific to the MG. An example would be a node that allows us to change the stiffness of the end-effector or specify the force applied by the end-effector. This means not only we can control the flow of execution but also specify controller specific values. This gives an additional control over the actual execution of a task.

A BTMG is a parameterized policy representation with parameters broadly categorized into two types: \textit{intrinsic} and \textit{extrinsic}, as mentioned in~\cite{ahmad2022generalizing,Ahmad2023LearningTA}.
\textit{Intrinsic} parameters encompass elements like the structure of BT, the quantity of BT nodes, and the type of motion generator employed. Conversely, \textit{extrinsic} parameters include variables such as the applied force, position offsets, and the end-effector's velocity. The specification of \textit{extrinsic} parameters can be done manually~\cite{rovida18btmg}, through reasoning, or by employing RL~\cite{mayr21iros,mayr2022skill,Ahmad2023LearningTA}, offering flexibility in adapting the BTMG framework to diverse tasks and environments. 

In our prior work, we have successfully used BTMG policy representation~\cite{rovida18btmg} for skill execution strategies in complex robotic tasks, including peg-in-a-hole, object pushing, and obstacle avoidance~\cite{mayr2022combining,mayr21iros}. We have further enhanced this approach by employing RL to dynamically learn and adjust the BTMG parameters~\cite{mayr2022combining}, allowing for adaptability in response to task variations~\cite{ahmad2022generalizing,ahmad2023learning}. Additionally, we have utilized planning techniques within the BTMG framework to sequence skills effectively~\cite{mayr2022skill} and demonstrated how incorporating priors can expedite the learning process~\cite{mayr22priors}.
\subsection{Learning parameters of BTMG}
\label{Learning parameters of BTMG}
In addition to predefined configurations, the extrinsic parameters of the BTMG representation can also be learned. This capability is crucial for adjusting the robotic skills dynamically, ensuring efficient task execution across diverse scenarios. To achieve this, we employ a policy optimization search method, as outlined in~\cite{chatzilygeroudis172iicirsi, chatzilygeroudis2019survey}, which is instrumental in learning the extrinsic parameters present in the \textit{action} nodes of the tree. The objective is to derive a policy $\pi$, where the action $\mathbf{u} = \pi(\mathbf{x}|\boldsymbol{\theta})$ is determined based on the state $\mathbf{x}$ and policy parameters $\boldsymbol{\theta}$, aimed at maximizing the expected long-term reward over $T$ time steps of policy execution. The optimization of these parameters is facilitated through \textit{Bayesian Optimization}~(BO)~\cite{nardi18hypermapper}, enabling the identification of extrinsic parameters that are adaptable to diverse situations. For an in-depth exploration of this optimization process and its applications, refer to~\cite{mayr2022skill, mayr21iros}.

\section{Approach}
\begin{figure}
    \centering
        \includegraphics[width=\columnwidth]{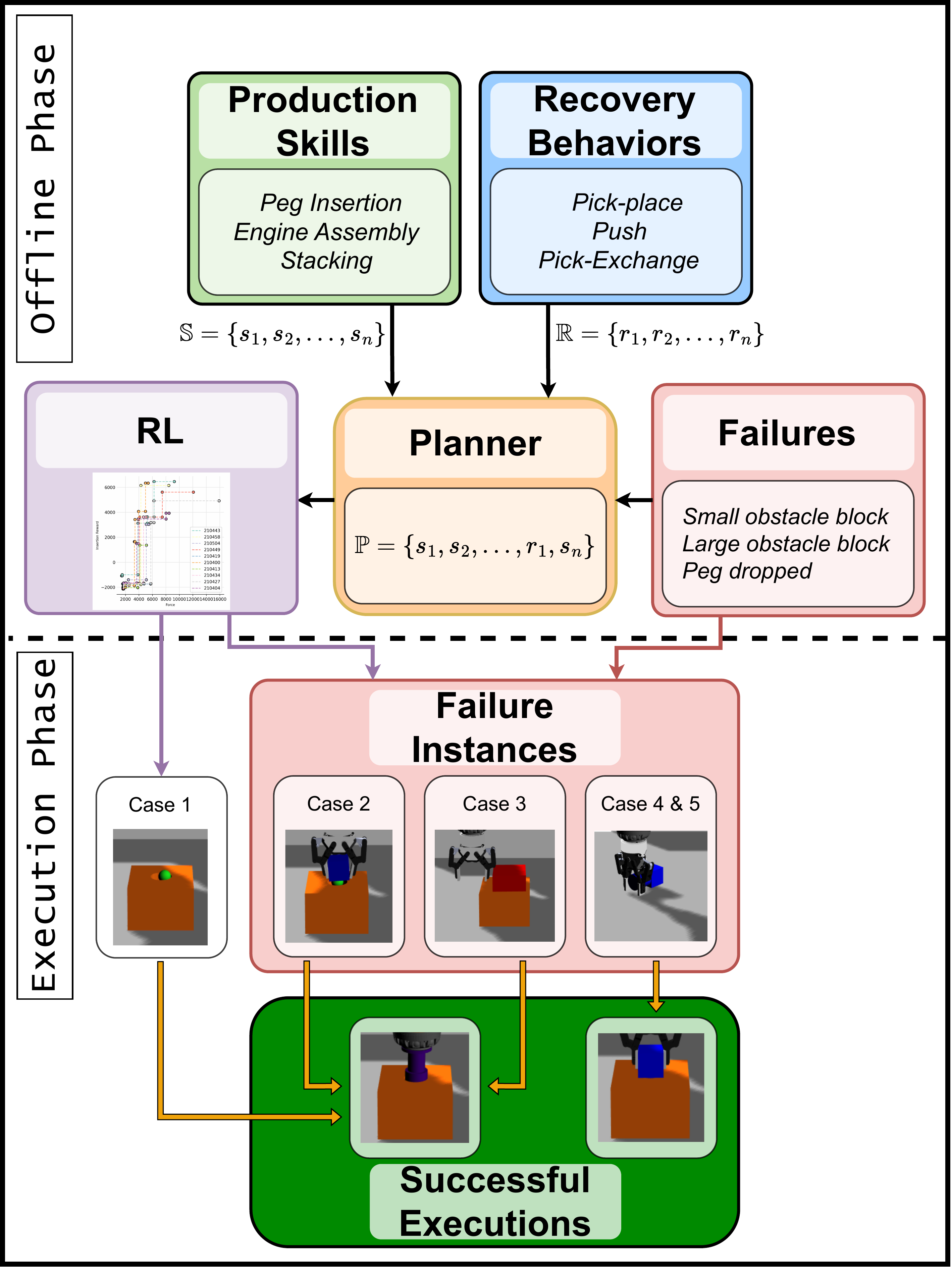}
        \caption{The figure shows the peg-in-a-hole task execution using our approach. We have separate sets of production and recovery behaviors. We can use a planner to come up with a sequence for a given failure specification. Subsequently, we tune the learnable parameters via reinforcement learning. Ultimately, appropriate recovery behaviors and skills are applied based on the identified failure, ensuring successful peg insertion. In the image we see successful peg insertions for all the scenarios.}
    \label{fig:Approach}
    \vspace{-0.1cm}
\end{figure}

In this section, we outline our approach, beginning with the assumptions guiding our work. We then detail the recovery behaviors implemented, the role of the planner in our framework, and conclude with an overview of the experimental scenarios designed to test our methodology.
The overall approach is shown in Figure~\ref{fig:Approach}.

\subsection{Assumptions}
Following are the assumptions for this work:
\begin{enumerate}
\item We operate under the premise that we are addressing expected and known failures within operational processes, leveraging human experience and historical data to anticipate these failures.
\item Given a comprehensive set of skill primitives, we assume our system has the capability to dynamically generate suitable recovery behaviors for any known failure, assuming a solution exists within the parameter space defined by these skills. 
\end{enumerate}
The first assumption acknowledges the predictability of common failures in operational processes, with an understanding that  basic parameters like object type, size, and weight are known and can be stored as knowledge about the system. For instance, in SkiROS2, this information can be managed using the world model~\cite{mayr2023skiros2, rovida2017skiros, mayr2023using}.

Addressing the second assumption more deeply, our approach leverages the natural capabilities of skills, planning, and parameter estimation to generate recovery behaviors tailored to specific failure scenarios. This method allows for the dynamic creation of recovery strategies by learning the necessary parameters (categorized as \textit{extrinsic} parameters within the BTMG framework) and determining the optimal sequence of skills for complex error situations. This strategy, inspired by the methodologies demonstrated in~\cite{styrud2023bebop}, ensures that we are not constrained to having a predefined skill for each recovery situation but can adapt and respond effectively to a wide range of failures.

\subsection{Recovery Behaviors}
In our framework, recovery behaviors are defined as specialized skills aimed at restoring a robotic system to its desired state after encountering a failure. These behaviors, defined by specific parameters, preconditions, and postconditions, are crafted from a predefined set of skill primitives, ensuring a seamless integration into the robot's comprehensive skill set. Within the BTMG policy representation, these recovery behaviors are integrated into the broader execution strategy, leveraging the capabilities of the SkiROS2 platform~\cite{rovida2017skiros} for effective implementation.

The generation of recovery behaviors from skill primitives is inspired by the approach outlined in~\cite{styrud2023bebop, nasiriany2022augmenting}, emphasizing the versatility and power of a well-defined set of primitives. Our set includes:
\begin{itemize}
\item \textbf{GripperOpen} and \textbf{GripperClose}, controlling the state of the gripper.
\item \textbf{GoToLinear}, moving the end-effector linearly while maintaining orientation. Also allows positional offsets in specified directions.
\item \textbf{ChangeStiffness}, adjusting the stiffness of the end-effector.
\item \textbf{ApplyForce}, applying force in a specified direction.
\end{itemize}

These primitives serve as the building blocks for constructing the recovery behaviors necessary for addressing specific failure scenarios encountered during task execution. The parameters for these behaviors can be finely tuned manually, through reasoning, or using RL, with RL playing a pivotal role in enhancing their robustness and adaptability. We identify which parameters require optimization through RL~\cite{mayr2022skill}, and refine them based on the task's needs using Bayesian Optimization (BO), as detailed in Section~\ref{Learning parameters of BTMG}. 
\begin{figure*}[!t]
    \centering
    \includegraphics[width=\textwidth]{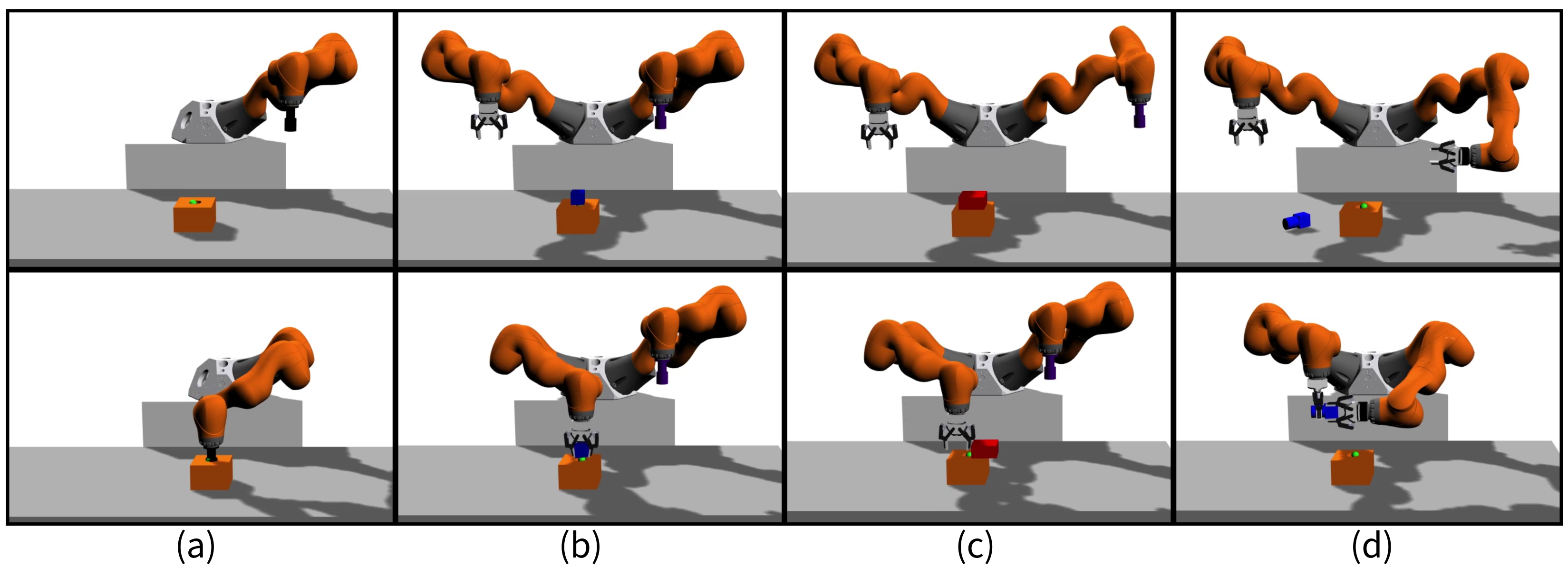}
    \caption{Illustration of the \textit{PegInsertion} skill alongside its associated failure scenarios and corresponding recovery behaviors. Panel (a) illustrates the initial and final states of Scenario~1. Panel (b) displays the initial state of Scenario~2 with the recovery behavior of \textit{pick-place}. Panel (c) showcases the initial state of Scenario~3 with the recovery behavior of \textit{push}. Lastly, panel (d) presents the initial states of Scenarios~4 and~5 with the recovery behavior of \textit{pick-exchange}.}
    \label{fig:scenarios}
    \vspace{-0.5cm}
\end{figure*}
Following are the recovery behaviors used in this paper:
\begin{itemize}
\item \textbf{Pick-Place:} Generated from \textit{GripperOpen}, \textit{GripperClose}, \textit{GoToLinear}, and \textit{ChangeStiffness}, this behavior enables obstacle removal, specifying parameters like \textit{arm} and \textit{obstacle}.
\item \textbf{Push:} Incorporating \textit{ApplyForce} alongside the other primitives, this behavior is tailored for displacing heavier obstacles, with parameters such as \textit{force} being optimized through RL.
\item \textbf{Pick-Exchange:} A complex behavior utilizing all five primitives to facilitate object transfer between arms. Additionally allows \textit{offsets} in x and y directions that can be set manually or through RL.
\end{itemize}
This methodology underscores the power of our set of skill primitives to generate the necessary recovery behaviors effectively. It also illustrates the ease with which this set can be extended should the need arise, ensuring that recovery behaviors are both situationally dependent and swiftly generated based on the available primitives. The adaptability and quick generation of these behaviors, as demonstrated in~\cite{styrud2023bebop}, are critical for our approach, allowing for rapid response to a wide range of failure scenarios.

\subsection{Planner}
In the context of collaborative tasks, our approach leverages the knowledge of expected failures to inform the design of recovery behaviors, encoding these failures as preconditions and postconditions. For instance, during a peg-in-a-hole task, a typical failure such as an obstruction in the hole by a small block can be explicitly defined as a precondition for triggering a recovery behavior. The successful clearance of the obstruction, resulting in an unblocked hole, is set as the postcondition, marking the completion of the recovery behavior. This structured encoding allows for the potential use of planning algorithms to sequence the necessary recovery behaviors for the task at hand, although it's important to note the practical challenges involved.

While we have previously demonstrated the use of the Problem Domain Description Language (PDDL) planner within the SkiROS2 framework to orchestrate sequences of skills~\cite{mayr2022skill} for robotic tasks, applying such planning in real-time collaborative scenarios poses unique challenges. In these settings, cycle times are critical, and it may not be feasible to invoke planning for every action, especially for tasks requiring rapid execution. However, a strategy~\cite{colledanchise2018behavior,styrud2023bebop} can be employed where preconditions are continuously monitored, and the planner is triggered only when specific conditions are not met, necessitating recovery actions. This approach ensures that planning is utilized efficiently, only when necessary to address deviations from expected task execution, thus maintaining operational efficiency while still benefiting from the adaptability offered by recovery behaviors. It is through this nuanced application of planning, tailored to the dynamics of collaborative tasks, that we can navigate the complexities of integrating automated recovery strategies effectively.

\subsection{Scenarios}
To evaluate our approach, we introduce a series of progressively more challenging scenarios within the peg-in-a-hole task, each necessitating distinct recovery behaviors, see Figure~\ref{fig:scenarios}. In every scenario, we focus on learning the parameters of the peg insertion skill. The differentiation among these scenarios hinges on the utilization of recovery behaviors, the method of parameterization for these behaviors (learned via RL or manually specified), and whether the execution of a recovery behavior necessitates a relearning of the \textit{PegInsertion} skill parameters due to changes in the task environment. In all the scenarios, we learn the parameters \textit{PegInsertion} skill via RL. 

\begin{enumerate}
\item \textbf{Baseline:} Utilizes only the \textit{PegInsertion} skill, serving as control.
\item \textbf{Static Recovery:} Uses a manually specified \textit{pick-place} recovery behavior to address a simple obstruction.
\item \textbf{Dynamic Recovery:} Employs a \textit{push} recovery behavior with RL-determined parameters for handling a more complex obstruction.
\item \textbf{Static Recovery with Behavior Changes:} Features a manually specified \textit{pick-exchange} recovery behavior that alters the task environment, necessitating the relearning of peg insertion skill parameters.
\item \textbf{Dynamic Recovery with Behavior Changes:} Similar to the previous scenario but uses RL to learn the \textit{offsets} for grasping the peg during the \textit{pick-exchange} recovery behavior.
\end{enumerate}


\begin{figure}[tpb!!]
	{
            \begin{center}
		      \includegraphics[width=0.8\columnwidth]{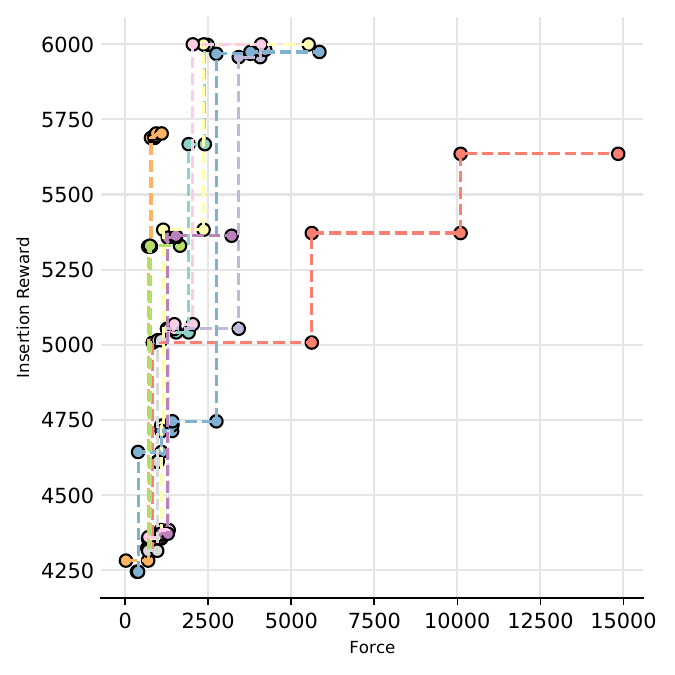}
            \end{center}
	}
    \vspace{-0.5cm}
	\caption{Pareto front for Scenario~1: \textit{Baseline}. Each experiment is denoted by a distinct color, with each bold point representing a pareto-optimal policy ready for execution. The optimizer tries to strike a balance between the reward for successful insertion and the force applied by the end-effector.}
	\label{fig:scenario1}
\end{figure}

\begin{figure}[tpb!!]
	{
            \begin{center}
		      \includegraphics[width=0.8\columnwidth]{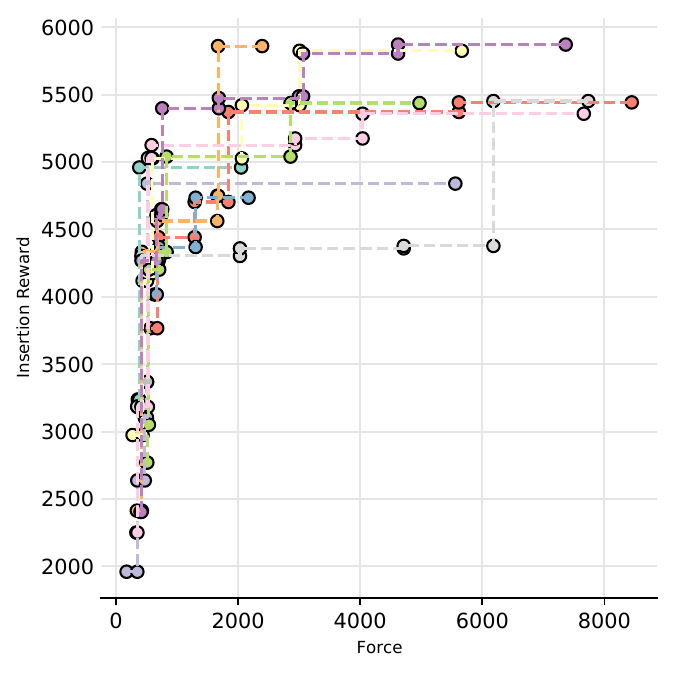}
            \end{center}
    }
    \vspace{-0.5cm}
	\caption{Pareto front for Scenario~2: \textit{Static Recovery}. This demonstrates that achieving a higher insertion reward necessitates greater force application, as observed from the force exerted by the end-effector during the search for the hole.}
	\label{fig:scenario2}
\end{figure}

\begin{figure}[tpb!!]
	{
            \begin{center}
		      \includegraphics[width=0.8\columnwidth]{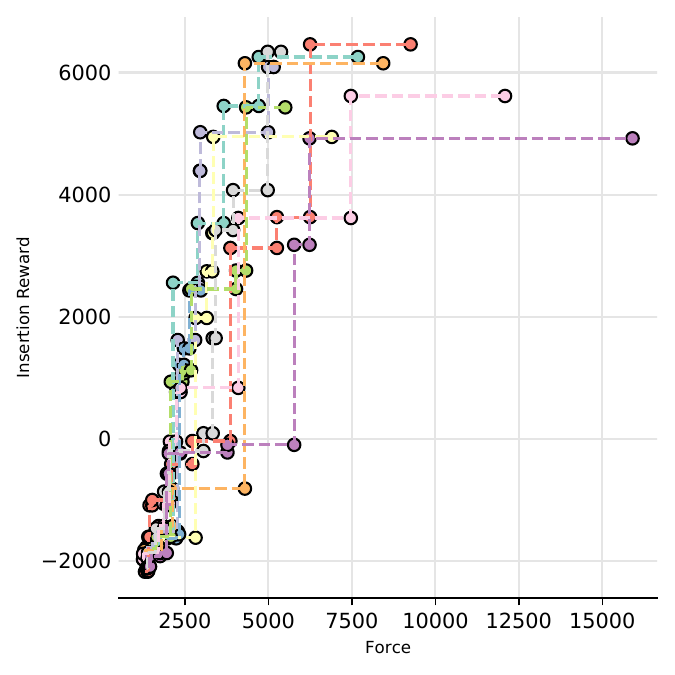}
            \end{center}
    }
    \vspace{-0.5cm}
	\caption{Pareto front for Scenario~3: \textit{Dynamic Recovery. In this scenario}, the application of force is notably higher because pushing the obstacle away requires additional force before the peg can be inserted into the hole.}
	\label{fig:scenario3}
\end{figure}

\begin{figure}[tpb!!]
	{
            \begin{center}
		      \includegraphics[width=0.8\columnwidth]{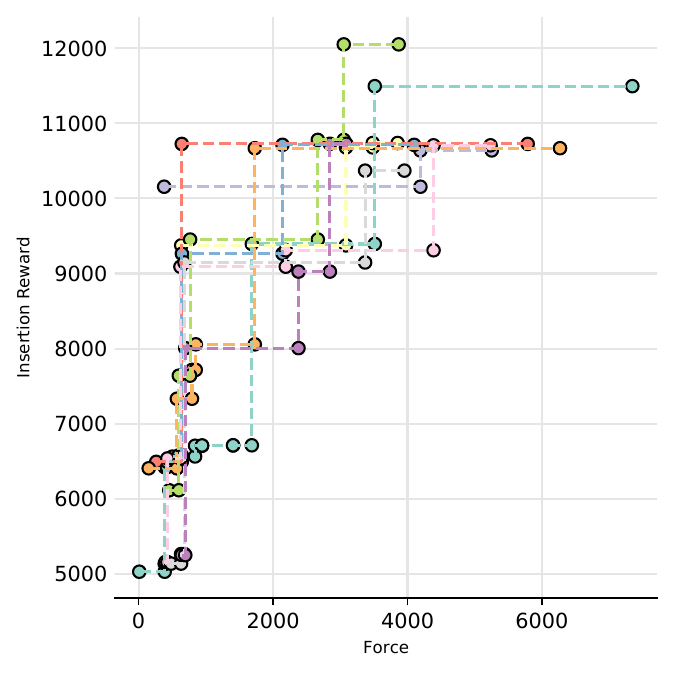}
            \end{center}
    }
    \vspace{-0.5cm}
	\caption{Pareto front for Scenario~4: \textit{Static Recovery with behavior changes}. In this scenario, we observe a similar trend to Scenario 2, where a higher force is necessary as the peg searches for the hole, reflecting the similar nature of the tasks.}
	\label{fig:scenario4}
\end{figure}

\begin{figure}[tpb!!]
	{
            \begin{center}
		      \includegraphics[width=0.8\columnwidth]{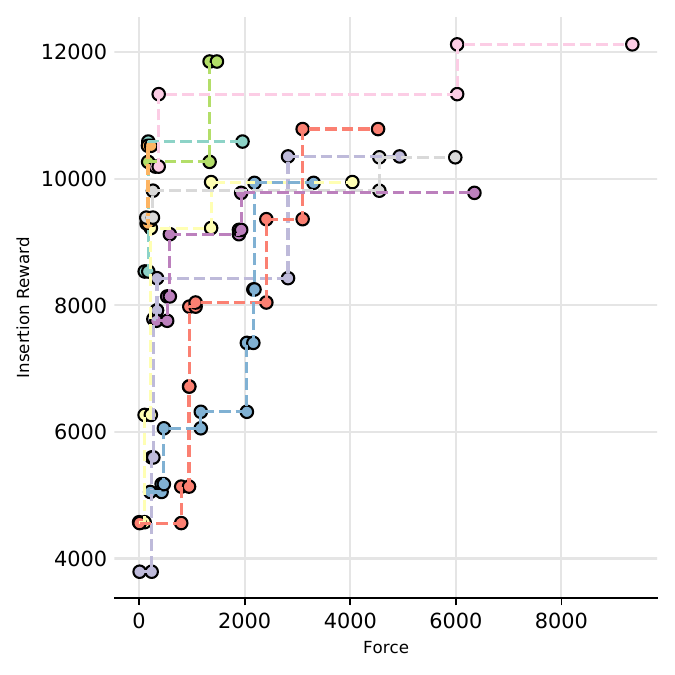}
            \end{center}
    }
    \vspace{-0.65cm}
	\caption{Pareto front for Scenario~5: \textit{Dynamic Recovery with behavior changes}. In this scenario, the wide distribution of policies suggests the task's complexity, as picking up and exchanging the dropped peg significantly impacts the success of the peg insertion.}
	\label{fig:scenario5}
\end{figure}

\section{Experimental Setup}
Our set of experiments evaluate the effectiveness of recovery behaviors in a peg-in-a-hole task, progressively introducing more challenging failures to require different recovery strategies. The task primarily utilizes a single production skill, the \textit{PegInsertion} skill, complemented by various implementations of \textit{pick-place}, \textit{push} and \textit{pick-exchange} recovery skills, differing based on the learning status of their parameters. The experiments are conducted in the \textit{DART} simulator~\cite{lee18joss}, employing a Cartesian impedance controller for arm manipulation~\cite{mayr2022c++}.

\subsection{Peg-in-a-hole Task}
The objective of our task is to insert a peg into a hole within a box, as depicted in Figure~\ref{fig:Robot-Setup} and \ref{fig:scenarios}. We utilize the \textit{GoToLinear} skill for precise end-effector positioning and the \textit{PegInsertion} skill for insertion. The \textit{PegInsertion} skill activates upon the end effector's arrival at the box's approach pose, where it dynamically adjusts the end effector's stiffness to zero in the z-direction~(downwards) and applies a targeted force in the same direction. Additionally, it incorporates an overlaying circular motion, akin to an Archimedean spiral, aimed at the box's center. The learnable parameters of this skill, including the end-effector's applied force, path velocity, path distance, and radius, are crucial for successful insertion. For an in-depth exploration of the BTMG representation and further skill specifics, we refer the interested reader to~\cite{mayr2022skill}.

The task is framed as a multi-objective challenge focusing on successful insertion and minimizing applied force, with reward metrics aligned with those in~\cite{mayr2022skill}. To evaluate the successful insertion, we employ a trio of reward metrics: the success of the BT execution, the proximity of the peg to the hole, and its distance from the box. For gauging the applied force, a singular reward metric quantifies the cumulative force exerted by the peg. To enhance system robustness, we use domain randomization, varying the location of the block with a hole by a standard deviation of \SI{8}{\milli\meter} via Gaussian distribution and changing the arm's starting position across five positions. For each scenario, we conduct 40 iterations, with each iteration being evaluated five times to account for domain randomization. Each scenario is repeated 10 times. This approach ensures robustness in our assessments by introducing variability in the task environment.  A policy is considered successful if it manages to achieve peg insertion in at least three out of the five evaluations. Across all scenarios, the clearance between the peg and the hole is maintained at \SI{3}{\milli\meter} to standardize the task difficulty.

\subsection{Results and Discussion}
Across all scenarios and for each repetition, we identified at least one policy capable of successfully inserting the peg into the hole, demonstrating the effectiveness of our recovery behaviors and the adaptability of the \textit{PegInsertion} skill under varied failures. Notably, the introduction of recovery behaviors, whether static or dynamic, did not impede the task's success, highlighting the robustness of our approach.
The Pareto fronts for each scenario illustrate the trade-off between insertion success and applied force, with diverse policies achieving the task across all repetitions, see Figure~\ref{fig:scenario1},~\ref{fig:scenario2},~\ref{fig:scenario3},~\ref{fig:scenario4} and~\ref{fig:scenario5}. This diversity underscores our approach's flexibility, enabling effective completion of a task subjected to multiple objectives.

By focusing on collaborative robots and leveraging the adaptability of recovery behaviors, our approach provides an alternative to automated recovery strategies. Even though, the static and dynamic scenarios we presented could be addressed  through automated recovery strategies, our method uses RL to dynamically adjust recovery behavior parameters, ensuring effective response to environmental changes. This adaptability, crucial for on-the-fly task management, sets our approach apart, offering a flexible solution to the changing demands of dynamic environments.

Additionally, it is pertinent to reference findings from our previous work~\cite{mayr2022skill}, where we evaluated the efficacy of using RL to specify parameters for the \textit{PegInsertion} skill against three distinct baselines: planning with predefined parameter values, random policy selection, and policies chosen by robot operators. The learned policies from our RL-based approach outperformed the alternative strategies in terms of success rates. Therefore, we opted not to directly compare our current approach with these baselines in the present study. Our focus in this study was the effectiveness of adaptable recovery behaviors for failure handling.
 Furthermore, it is also worth mentioning that the adaptability of our approach can be further enhanced by accommodating different task variations, as demonstrated in our previous work~\cite{Ahmad2023LearningTA}. In that study, we trained a model to predict the long-term reward of different policies, showing that the policies suggested by this model perform comparably to those optimized directly through RL. In principle, this predictive model could be used in place of direct RL optimization, potentially accelerating the adaptability process for recovery behaviors in response to varying task conditions.
\section{Conclusion and Future Work}
In this paper, we presented a novel approach that models recovery behaviors as robotic skills to effectively manage and recover from failures in robotic tasks. By defining these recovery behaviors with specific parameters, preconditions, and postconditions, and utilizing the BTMG policy representation as the execution strategy, we have demonstrated a structured method to represent and implement these behaviors. The adaptability of these behaviors is enhanced through RL to dynamically adjust parameters. Our approach enables robots to autonomously recover from disruptions and resume normal operations seamlessly.
We evaluated our methodology through the peg-in-a-hole task by gradually introducing challenging failures and recovering from them, thereby testing the resilience and adaptability of our recovery strategies. By framing this task as a multi-objective challenge, focusing on successful insertion while minimizing applied force, we showcased the effectiveness of our approach. Our results confirm that the integration of recovery behaviors, modeled as adaptable robotic skills within the BTMG framework, significantly enhances the robot's ability to recover from failures, thereby improving operational efficiency and task success rates.
    
In our future work, we aim to develop a comprehensive recovery pipeline that not only identifies failures but also selects the appropriate recovery skills automatically to address them effectively. This pipeline will enhance our current set of recovery skills, making them capable of handling not just anticipated failures but also unexpected ones within certain limits. Our goal is to leverage the structure of a BT and its tick signals, which return different states, to pinpoint the exact location of a failure by analyzing which node returns a failure state. This diagnostic capability will enable us to match the specific pre- and post-conditions of a failure, facilitating the selection of suitable recovery behaviors from a more generalized skill set. To achieve this, we plan to explore the use of a recursive tree structure~\cite{colledanchise2018behavior}, which will play a crucial role in dynamically choosing the most effective recovery behavior based on the situation at hand. Additionally, we will also like to explore the creation of a dataset of various failures to demonstrate learning recovery behaviors from skill primitives, akin to a reformulation in~\cite{styrud2023bebop} focused on error recovery. This effort will include verifying the sufficiency of our skill primitives for comprehensive recovery scenarios, enhancing the system's adaptability and resilience in complex environments.

\section*{ACKNOWLEDGMENTS}
We thank Momina Rizwan and Simon Kristoffersson Lind for the interesting discussions and the constructive feedback.
This work was partially supported by the Wallenberg AI, Autonomous Systems and Software Program (WASP) funded by Knut and Alice Wallenberg Foundation. We acknowledge that we have used Generative AI language tools for editing of the author's original text. This editing includes sentence structuring, spelling and grammar corrections.

\bibliography{2024-CASE}
\bibliographystyle{bib/IEEEtran}

\end{document}